\documentclass[letterpaper, 10 pt, conference]{ieeeconf}  

\IEEEoverridecommandlockouts                              

\usepackage{graphics} 
\usepackage{mathptmx} 
\usepackage{times} 
\usepackage{amsmath} 
\usepackage{amssymb,bm}  
\usepackage{mathptmx}
\usepackage{mathtools}
\usepackage{url}
\usepackage{tikz}
\usepackage[linesnumbered,ruled, vlined]{algorithm2e}
\usepackage[tight]{subfigure}
\usepackage{multirow}
\usepackage{wrapfig}

\newcommand{\Def}[1]{\textbf{\ttfamily\color{green!50!black}#1}\unskip}

\newcommand{\Key}[1]{\textbf{\ttfamily\color{blue!50!black}#1}\unskip}

\definecolor{commentcolor}{RGB}{110,154,155}   
\newcommand{\PyComment}[1]{\ttfamily\textcolor{commentcolor}{\# #1}}  
\newcommand{\PyCode}[1]{\ttfamily\textcolor{black}{#1}} 

\SetKwInput{KwInput}{Input}                
\SetKwInput{KwOutput}{Output}              

\DeclareMathAlphabet{\mathcal}{OMS}{cmsy}{m}{n}

\title{\LARGE \bf
IDA: Informed Domain Adaptive Semantic Segmentation
}

\author{Zheng Chen$^{1}$, Zhengming Ding$^{2}$, Jason M. Gregory$^{3}$, and Lantao Liu$^{1}$
\thanks{\newline
$^{1}$Z. Chen and L. Liu are with Luddy School of Informatics, Computing, and Engineering, Indiana University, Bloomington, IN 47408, USA. Email: {\tt\small \{zc11, lantao\}@iu.edu} 
\newline
$^{2}$Z. Ding is with the Department of Computer Science at Tulane University. Email: 
{\tt\small zding1@tulane.edu}
\newline
$^{3}$J. Gregory is with DEVCOM Army Research Laboratory, USA.
}
}

\begin{document}

\maketitle
\thispagestyle{empty}
\pagestyle{empty}

\begin{abstract}
Mixup-based data augmentation has been validated to be a critical stage in the self-training framework for unsupervised domain adaptive semantic segmentation (UDA-SS), which aims to transfer knowledge from a well-annotated (source) domain to an unlabeled (target) domain. 
Existing self-training methods usually adopt the popular region-based mixup techniques with a random sampling strategy, 
which unfortunately ignores the dynamic evolution of different semantics across various domains as training proceeds. 
To improve the UDA-SS performance, we propose an \textbf{I}nformed \textbf{D}omain \textbf{A}daptation (IDA) model, a self-training framework that mixes the data based on class-level segmentation performance, which aims to emphasize small-region semantics during mixup. In our IDA model, the class-level performance is tracked by an expected confidence score (ECS). We then use a dynamic schedule to determine the mixing ratio for data in different domains. 
Extensive experimental results reveal that our proposed method is able to outperform the state-of-the-art UDA-SS method by a margin of 1.1 mIoU in the adaptation of GTA-V to Cityscapes and of 0.9 mIoU in the adaptation of SYNTHIA to Cityscapes. Code link: {\color{purple}\url{https://github.com/ArlenCHEN/IDA.git}}
\end{abstract}

\section{Introduction}
\label{sec:intro}
\begin{figure} [t]
{
\centering
  {\includegraphics[width=\linewidth]{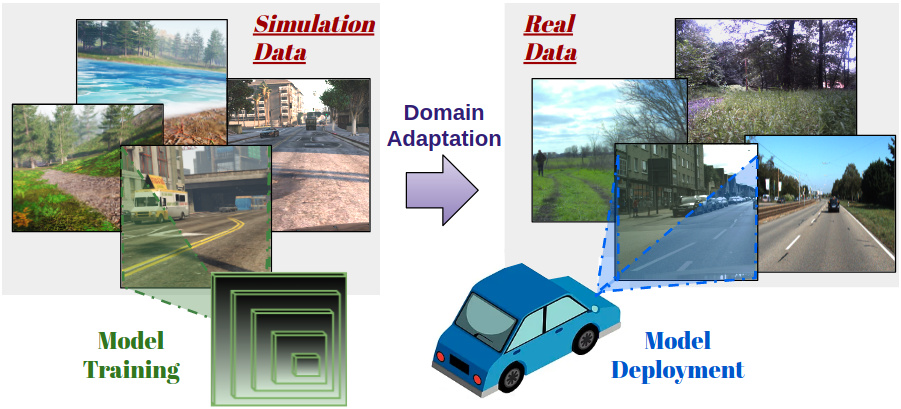}}
\caption{\small Domain adaptation aims at diminishing the model performance drop due to the shift of different data domains, e.g., simulated data vs. real data.
} 
\label{fig:intro}  \vspace{-15pt}
}
\end{figure}

Semantic segmentation (SS) aims to learn a pixel-wise classification for a given image and plays a critical role in various applications such as infrastructure/industrial inspections \cite{rubio2019multi}, biomedical diagnoses \cite{ronneberger2015u}, and vehicle autonomy \cite{richter2016playing}. 
Current mainstream segmentation models \cite{xie2021segformer} \cite{zhang2021k} \cite{strudel2021segmenter} heavily rely on deep neural networks (DNNs) which usually require a huge amount of manual annotations in order to achieve desirable performance, e.g., labeling for a single image might require more than $1.5$ hours on average \cite{cordts2016cityscapes}. There exist some public datasets that provide dense annotations, e.g., Cityscapes \cite{cordts2016cityscapes}, ACDC \cite{SDV21}. However, these existing datasets are far from providing sufficient coverage for other miscellaneous novel environments, leading to deep models that fail to generalize. 
In this case, how to transfer the knowledge in the easy-to-access data (e.g., data in simulators; existing public datasets) to boost the model generalization for unseen data of other domains is critical as there is usually a domain shift between the data used to train the model and the data encountered during deployment (see Fig. \ref{fig:intro}).

To achieve the model transfer, a broadly studied task is \textit{domain adaptation} (DA), where we define the data with labels as a \textit{source} domain, and the data to be processed and used for prediction as a \textit{target} domain. In many real-world scenarios, it can be challenging or costly to quickly create labels for target data. In this case, {\em unsupervised domain adaptation} (UDA) provides a means to solve the special DA setting where the target domain has no labels. 

A critical step for UDA-SS is the mixup-based data augmentation which oftentimes utilizes random sampling  
where certain rectangular regions (e.g., in CutMix \cite{yun2019cutmix}) or class regions (e.g., in ClassMix \cite{olsson2021classmix}) 
are randomly selected with a predefined size and mixed. 
This strategy can generally mix the visual contents and mitigate the understanding shift of the DNNs. 
Unfortunately, this strategy ignores the dynamic evolution of the two domain data along the training progress, 
and can fail in leveraging important ``informative'' data that however are obscured amid the process. 
The ``informativeness" of data can be defined as its significance to the final performance. 
In greater detail, in UDA-SS the segmentation of multiple classes are predicted such that the final performance of the model depends on the performance of each individual class. We conjecture that the model performance is mainly driven by the performance of ``bottleneck'' classes that usually have low-quality segmentations. 
Note also, the bottleneck classes may vary during the training process. 
The model performance can be significantly improved if we can identify what bottleneck classes are so as to inform us to specially improve their performance. 
Regions associated with those bottleneck classes are defined as informative data, and our proposed method is thus termed as 
Informed Domain Adaptation (IDA). 

The key to the proposed IDA framework is a novel mixup technique --- {\em Informed Mix} (IMix) built upon ClassMix~\cite{olsson2021classmix}. 
Different from the ClassMix, our IMix bridges data from two domains according to the confidence values of training progress indicators. 
Thus the IMix is informed by and prioritizes, the regions indicated by a low indicator value which means the classes are balanced in both raw image space and label space, and the training will be unbiased, leading to an increase of final performance.
To comply with the training dynamics, we also propose a novel adaptation schedule for IDA. The proposed schedule adaptively determines the ratio of image regions from two domains for mixing because the dynamic changes in training progress are related not only to the type of bottleneck classes but also to the number of bottleneck classes. 
Setting a fixed number will either miss effective data or introduce possible noise. 
Our proposed schedule has three phases. 
In the first phase, IMix mainly selects easy classes from source images for adaptation. In the second phase, hard/bottleneck classes from source images are selected and the number of the selected classes decreases from a high level to a low level while the selections from target images increase from a low level to a high level. In the third phase, the numbers of selections from the source and target domains will be maintained at low and high levels, respectively. To summarize the contribution of this work,
\begin{itemize}
    \item
    We propose a principled model, IDA, for the UDA-SS. The IDA model is a self-training framework that exploits the obscured informativeness of data, which has not been previously studied in DA.
    \item 
    We propose a new mixup technique, IMix, that bridges the source and target domains according to the training progress defined by an expected confidence assessment.
    \item
    We propose a novel dynamic adaptation schedule which can adaptively adjust the mixing ratio for different domains to optimize the adaptation efficiency. We will make the code of this work public.
\end{itemize}

Finally, our extensive evaluations on popular datasets show that our IDA outperforms the current SOTA model HRDA \cite{hoyer2022hrda} with a remarkable margin under the same settings.

\section{Related Work}
\label{sec:related}
\textbf{Domain Adaptive Semantic Segmentation:}
Mainstream methods for tackling UDA can be categorized into two classes --- feature alignment (FA) \cite{ganin2016domain} \cite{hoffman2016fcns} \cite{hoffman2018cycada} \cite{tsai2018learning} \cite{saito2018maximum} \cite{vu2019advent} \cite{luo2019taking} \cite{chen2022cali} \cite{wang2020classes} and self-training (ST) \cite{zou2018unsupervised} \cite{zhang2017curriculum} \cite{mei2020instance} \cite{hoyer2022daformer} \cite{hoyer2022hrda}. FA adapts the model by aligning the features from the source domain to the target domain using adversarial training, i.e., features from two domains are expected to be indistinguishable through a domain discriminator.
However, FA suffers from two issues. First, FA aligns features from two domains in a \textit{global} way by evaluating the domain discrimination using features of the whole image. This can be problematic for semantic segmentation as each image contains multiple classes. Aligning features globally cannot guarantee the class-level shift is eliminated, and even worse,  it is possible that features are aligned at a global level but severely misaligned at a class level, causing the so-called {negative transfer} \cite{zhang2020survey}. Recently, some class-level FA methods \cite{saito2018maximum} \cite{luo2019taking} \cite{wang2020classes} \cite{chen2022cali} are proposed to consider a finer level of feature structure, but they still suffer from the lack of target labels and show a weak performance. Second, FA adopts an adversarial training paradigm which is known to be unstable to train \cite{goodfellow2020generative}. On the other hand, ST tackles the UDA by a teacher-student framework\cite{hoyer2022hrda} \cite{hoyer2022daformer} \cite{araslanov2021self}, where the teacher is trained on the source domain and predicts \textit{pseudo labels} for target images. Then the student is supervised by those predicted pseudo labels. Recently ST \cite{hoyer2022hrda} \cite{hoyer2022daformer} has been prevalent since it constantly breaks the state-of-the-art record of UDA-SS due to the highly efficient feedback for adaptation from pseudo-labels.

\textbf{Mixup Data Augmentation:}
mixup-based data augmentation has been demonstrated to be a vital step for UDA-SS as it can achieve adaptation directly in the raw input space and label space, by forcibly mixing different domains for each data sample. Region-based mixups, e.g., CutMix \cite{yun2019cutmix} and ClassMix \cite{olsson2021classmix} are two representative mixups used in UDA-SS. CutMix and ClassMix typically adopt a random sampling strategy when mixing data from two domains, i.e., the region in one domain is randomly selected with a predefined size while the rest regions are from the other domain. DACS \cite{tranheden2021dacs} is recently proposed to apply the idea of ClassMix to domain adaptation. DACS randomly selects regions of half of the classes in source images and pastes the target data to the rest regions. DACS has been validated to be effective in recent ST methods \cite{hoyer2022daformer}, \cite{hoyer2022hrda}.

\vspace{-3pt}
\section{Methodology}\vspace{-3pt}
\label{sec:method}

To organize the presentation, 
in Sect.~\ref{sec:preliminaries}, we provide preliminary knowledge about the domain adaptive semantic segmentation. In Sect.  \ref{sec:overall_model}, we describe the general structure of our IDA model. In Sect.~\ref{sec:imix}, we first describe how we perform the identification of bottleneck classes on the fly along the training process. Then we introduce the IMix data augmentation by carefully considering the spatiotemporal changes of domain data during training.

\vspace{-3pt}
\subsection{Preliminaries of UDA-SS}\vspace{-3pt}
\label{sec:preliminaries}

We consider a source domain distribution $\mathcal{S}$ and a target domain distribution $\mathcal{T}$ over the joint space of $\mathcal{X}\times \mathcal{Y}$, where $\mathcal{X}$ is the input space and $\mathcal{Y}$ is the label space, respectively. In UDA-SS, we have access to $N_s$ labeled samples $\left ( \mathcal{X}_s = \left\{ x_i^s \right\}_{i=1}^{N_s}, \mathcal{Y}_s = \left\{ y_i^s \right\}_{i=1}^{N_s} \right )$ for the source domain, and only access to $N_t$ raw images $\mathcal{X}_t = \left\{  x_j^t\right\}_{j=1}^{N_t}$ for the target domain. A neural network $g$ comprising of a feature extractor $f_{\theta}$ parameterized by $\theta$ and a segmentation head $h_{\phi}$ parameterized by $\phi$, i.e., $g_{\theta, \phi} = h_{\phi}(f_{\theta})$, is usually adopted as the adaptation model. The expected error on the source domain is denoted by
\begin{equation}
    \label{eq:source_error}
    L_{\mathcal{S}} (\theta, \phi) = \mathbb{E}_{(x, y)\sim \mathcal{S}}\left [ l(g_{\theta, \phi}(x), y) \right ],
\end{equation}
where $l(\cdot, \cdot)$ represents the loss function. In UDA-SS, typically the standard cross-entropy with one-hot ground-truth (gt) label is used to compute the training loss:
   $ l(g_{\theta, \phi}(x), y) = -\sum_{c=1}^C \left [ y^c \cdot \log g_{\theta, \phi}(x)^c \right ] $,
where $C$ is the class size.

Similarly, the expected error on the target domain is denoted by $L_{\mathcal{T}} (\theta, \phi)$, but we cannot obtain an expression for $L_{\mathcal{T}} (\theta, \phi)$ as we have no labels for target data. However, we have an indirect way to approximate $L_{\mathcal{T}} (\theta, \phi)$ by 
\begin{equation}
\label{eq:target_error}
    L_{\mathcal{T}} (\theta, \phi) = \mathbb{E}_{x\sim \mathcal{T}}\left [ l(g_{\theta, \phi}(x), \hat{y}) \right ],
\end{equation}
where $\hat{y}$ is a pseudo label generated by the model trained on the source domain, $\hat{y} = \texttt{one-hot}(\texttt{argmax}_c(g_{\theta_s, \phi_s}(x)))$, where $\theta_s$ and $\phi_s$ represent the neural network parameters trained on the source domain.
Based on Eq. (\ref{eq:source_error}) and Eq. (\ref{eq:target_error}), we can obtain the adaptation objective as
\begin{equation}
\label{eq:adaptation_objective}
    \min_{\theta, \phi} L_{\mathcal{S}}(\theta, \phi) + L_{\mathcal{T}}(\theta, \phi).
\end{equation}

\vspace{-3pt}
\subsection{Model Framework}\vspace{-3pt}
\label{sec:overall_model}

\begin{figure} \vspace{-8pt}
  \centering
   \includegraphics[width=0.8\linewidth]{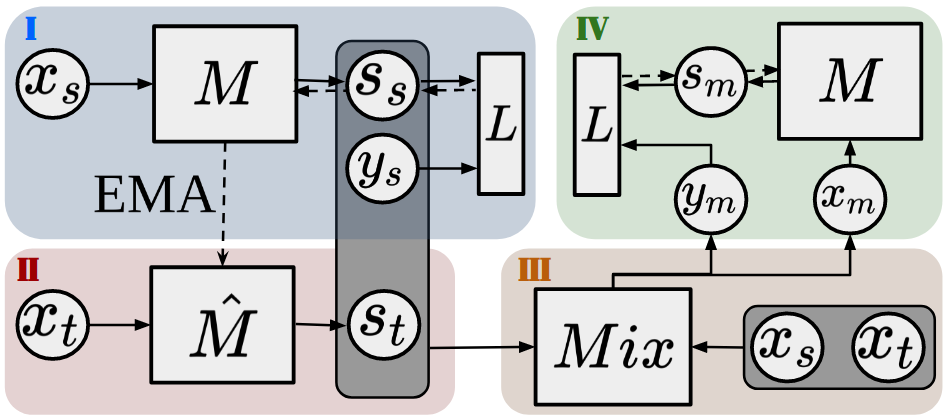}\vspace{-8pt}
   \caption{\small Overall structure of our IDA model. Four training stages are involved. $x$ represents the input image; $y$ denotes the label; $s$ is the model prediction; $L$ means the standard cross-entropy loss; Model $M$ is the student model while $\hat{M}$ is the teacher model.
   \vspace{-10pt}
   }
   \label{fig:model-overview}
\end{figure}

We build the proposed IDA model based on the teacher-student model in self-training (see Fig. \ref{fig:model-overview}, $M$ represents the student model and $\hat{M}$ represents the teacher model). 
The teacher model shares the same network structure as the student's. Four stages are involved in one training iteration. 
In the beginning of each iteration, we use the 
{\em exponential moving average} (EMA) to update the teacher model's parameters by the ones of the student model such that the teacher model can be synchronized with the latest weights of the student model. Then the student model 
is trained with source labels. In the second stage, we use the teacher model to generate pseudo labels for target images without back-propagation. In the third stage, our proposed IMix module (described in Sect.~\ref{sec:imix} later) takes as input the source image, source prediction, source label, target image, and target prediction (pseudo label) to generate a new pair of data that mixes the data from the two domains. In the fourth stage, the student model is further trained using the newly generated mixed data.

In the proposed IDA model, our approximation to the expected error on the target domain differs from the Eq. (\ref{eq:target_error}), where the generated pseudo label $\hat{y}$ is directly used for training the model. Instead, we use the newly generated data pair ($x_m$, $y_m$) to compute $L_{\mathcal{T}}$, as illustrated in Fig.~\ref{fig:model-overview}. 
We denote the distribution of the mixed data as $\mathcal{M}$. The adaptation objective of our IDA model is
\begin{equation}
    \label{eq:IDA_objective}
    \min_{\theta, \phi} L_{\mathcal{S}}(\theta, \phi) + L_{\mathcal{M}}(\theta, \phi).
\end{equation}
Using $L_{\mathcal{M}}$ to approximate $L_{\mathcal{T}}$ has been validated in existing work \cite{tranheden2021dacs} \cite{hoyer2022daformer} \cite{hoyer2022hrda}. The reason for this effectiveness lies in that Eq. (\ref{eq:adaptation_objective}) separates the supervision from the source domain and the target domain while Eq. (\ref{eq:IDA_objective}) mixes supervision signals from the two domains in the term of $L_{\mathcal{M}}$. This mixture can efficiently guide the model to understand the target data with the accompanying source data at the sample level.

\subsection{Informed Mix}
\label{sec:imix}

We propose Informed Mix (IMix) which is an important module in our IDA.
The IMix considers dynamic temporal-spatial changes of data during the training process and is able to adjust the mixture strategy accordingly. Different from previous mixup techniques \cite{yun2019cutmix} \cite{olsson2021classmix} \cite{tranheden2021dacs} that use random sampling to select mix regions from images, our IMix informatively selects the regions based on the class-level performance during training. 

\begin{figure} 
{
  \centering
    \subfigure[]  	      {\label{fig:ce_2}\includegraphics[width=0.24\linewidth]{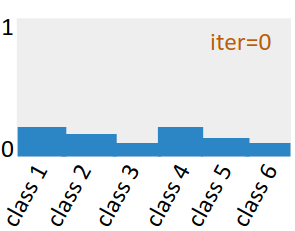}}
    \subfigure[]{\label{fig:ce_3}\includegraphics[width=0.24\linewidth]{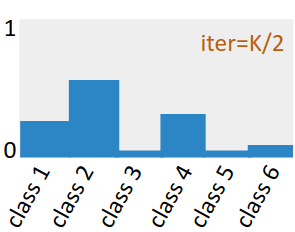}} 
    \subfigure[]{\label{fig:ce_4}\includegraphics[width=0.24\linewidth]{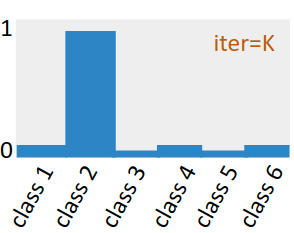}}
    \subfigure[]{\label{fig:one-hot}\includegraphics[width=0.24\linewidth]{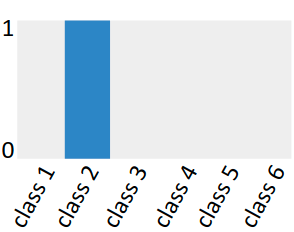}}
  \caption{\small (a) - (c) The changes of the categorical probability under the supervision of a (d) one-hot vector in different training iterations.  $K$ represents the total number of training iterations.
  }
\label{fig:ce}  
}
\end{figure}

\subsubsection{Class-level Performance Indicator}\vspace{-3pt}
\label{sec:indicator}

\begin{figure}[t]
  \centering
   \includegraphics[width=0.98\linewidth]{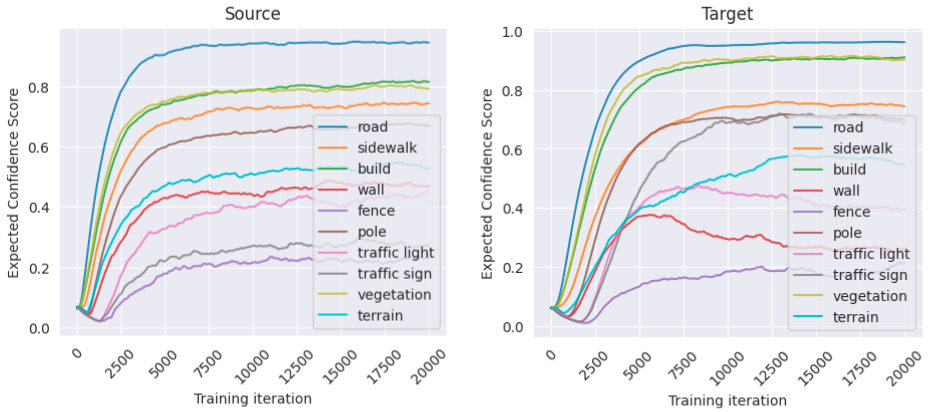}\vspace{-10pt}
   \caption{\small Smoothed ECS values for source  and target classes.} \vspace{-5mm}
   \label{fig:domain_ecs}
\end{figure}

To monitor the training progress,
we usually observe how the loss changes over multiple epochs/iterations. However, we have no labels for target data in UDA-SS and thus are unable to use a loss-like indicator. In this work, we propose to use the {\em confidence score} (CS) of the predicted probability as the indicator.  
CS has been used by previous works as the uncertainty approximation, e.g., \cite{zou2019confidence}.
The metric is defined as $\texttt{CS}(x) = \texttt{max} (g_{\theta, \phi}(x))$, where the model $g$ usually has the \texttt{softmax} function as the last layer. 
The reason for using the confidence score is related to the standard cross-entropy (CE) loss function which assumes that the label is a one-hot vector and minimizing the CE loss is equivalent to maximizing the CS. In this case, a higher CS value can indicate a lower loss and thus better performance. 

A simple illustration can be seen in Fig. \ref{fig:ce}, where Fig. \ref{fig:one-hot} shows a one-hot vector label, the change of the corresponding probability over multiple training iterations can be seen from Fig. \ref{fig:ce_2} to Fig. \ref{fig:ce_4}. The increased CE value can be a dual form of training error for indicating the training progress. 
By tracking the CS value during training, we are able to monitor the class-level performance on the fly, thus we can adaptively identify the data of the maximal informativeness.

In this work, we use the {\em expected confidence score (ECS)} as the class-level performance indicator. The ECS for class $c$ can be computed by
\begin{equation}
    \label{eq:ecs}
    \texttt{ECS}_c (x) = \mathbb{E}_{x\sim c} \left [ \texttt{CS}(x) \right ],
\end{equation}
where with a slight abuse of notation, we use the first $c$ to conceptually represent the $c^{th}$ class, while the second $c$ to represent the \textit{distribution} of the $c^{th}$ class. The ECS for the source domain and the target domain can be expressed by
\begin{equation}
    \label{eq:ecs}
    \begin{aligned}
        &\texttt{ECS}_c^s (x) = \mathbb{E}_{x\sim \mathcal{S}, x\sim c} \left [ \texttt{CS}(x) \right ], \\
        &\texttt{ECS}_c^t (x) = \mathbb{E}_{x\sim \mathcal{T}, x\sim c} \left [ \texttt{CS}(x) \right ].
    \end{aligned}
\end{equation}

To further validate the property of $\texttt{ECS}_c^s$ and $\texttt{ECS}_c^t$, we show the changes of the two ECSs during training in Fig. \ref{fig:domain_ecs}. Here we smooth the value of raw ECS by the EMA.
\begin{equation}
    \label{eq:ema_ecs}
    \begin{aligned}
        &{}^{j}\texttt{ECS}_{c}^s \leftarrow \tau \cdot {}^{j-1}\texttt{ECS}_c^s + (1-\tau) \cdot {}^{j}\texttt{ECS}_c^s, \\
        &{}^{j}\texttt{ECS}_{c}^t \leftarrow \tau \cdot {}^{j-1}\texttt{ECS}_c^t + (1-\tau) \cdot {}^{j}\texttt{ECS}_c^t,
    \end{aligned}
\end{equation}
where $\tau$ is the smoothness weight; $j$ represents the $j^{th}$ iteration during training. As we can see from Fig. \ref{fig:domain_ecs}, ECS values of almost-all classes are monotonously increasing during training for both the source domain and target domain. 

\begin{figure}[t] 
  \centering
   \includegraphics[width=0.95\linewidth]{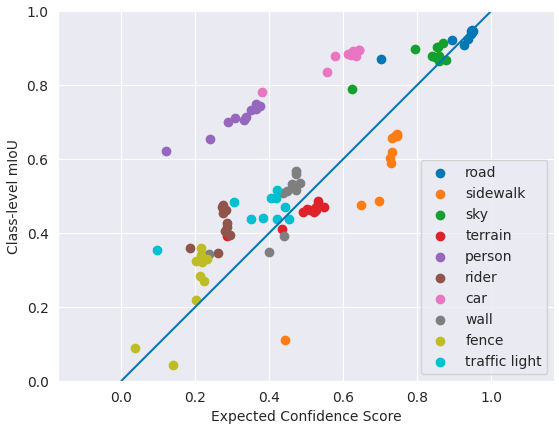} 
   \vspace{-10pt}\caption{\small Reliability diagram for class-level ECS. }\vspace{-10pt}
   \label{fig:reliability}
\end{figure}

To further validate that the class-level ECS is well calibrated, we also show the reliability diagram for some of the representative classes from Cityscapes \cite{cordts2016cityscapes} in Fig. \ref{fig:reliability} where we show the relation between the ECS values and the widely-adopted segmentation performance metric {\em mean-Intersection-over-Union (mIoU)}. We can see that in general the mIoU is positively correlated with the value of ECS.

\begin{algorithm}[h]
\SetAlgoLined
    \PyCode{\textbf{def}} \Def{ClassSample}($i_a$):\\
    \Indp
        \PyCode{$M$=\Key{zero\_like}($i_a$.shape)}\\
        \PyCode{$u_a$}=\Key{unique}($i_a$)\\
        \PyCode{$\hat{u}_a$=\Key{randint}(0, C, size=int(C/2))}\\
        \PyCode{$M$[i,j] = 1 if $i_a$[i,j] is in $\hat{u}_a$}\\
        \PyCode{\textbf{return} $M$}\\
    \Indm
    ~~\\
    \textbf{def} \Def{ISample}($i_a$, e, $\eta$):\\
    \Indp
        \PyCode{$M$=\Key{zero\_like}($i_a$.shape)}\\
        \PyCode{$u_a$}=\Key{unique}($i_a$)\\
        \PyCode{$k = \eta \cdot C$ }\\
        \PyCode{$t_a$ = \Key{topk}(e, k)}\\
        \PyCode{$\hat{u}_a$ = $t_a$.indices}\\
        \PyCode{$M$[i,j] = 1 if $i_a$[i,j] is in $\hat{u}_a$}\\
        \PyCode{\textbf{return} $M$}\\
    \Indm
    ~~\\
    \PyComment{$x_a,~x_b$: images from two domains}\\
    \PyComment{F is the segmentation model}\\
    \PyComment{C is the number of classes}\\
    \PyComment{e $\in \mathbb{R}^C$: ECS values for all classes}\\ 
    \PyComment{$\eta$: Allocation ratio}\\
    \textbf{def} \Def{Mix}($x_a$, $x_b$, method, e, $\eta$):\\
    \Indp
        \PyCode{$y_a$ = F($x_a$)} \PyComment{$y_a$ shape: [C, H, W]}\\
        \PyCode{$y_b$ = F($x_b$)} \\
        \PyCode{$i_a$ = \Key{argmax}($y_a$, axis=0)}\\
        \PyCode{if method == 'ClassMix':}\\ 
        \Indp
            $M$ = \Def{ClassSample}($i_a$) \PyComment{ClassMix}\\
        \Indm
        \PyCode{elif method == 'IMix':}\\
        \Indp
            $M$ = \Def{ISample}($i_a$, e, $\eta$) \PyComment{IMix}\\
        \Indm
        \PyCode{$x_m = M\odot x_a + (1- M)\odot x_b$}\\
        \PyCode{$y_m = M\odot y_a + (1-M)\odot y_b$}\\
        \PyCode{\textbf{return} $x_m$,$y_m$}\\
    \Indm
\caption{Functions of ClassMix and IMix} 
\label{algo:mix} 
\end{algorithm} 

\subsubsection{Class Selection Strategy}
\label{sec:class_selection}

We propose our {\em Informed Mix} (IMix) based on the DACS~\cite{tranheden2021dacs} which is a variant of ClassMix \cite{olsson2021classmix}, and build upon the idea of bridging images across domains.
DACS is the first to apply the idea of ClassMix \cite{olsson2021classmix} to domain adaptation. In DACS,  $x_a$ is selected from the source domain while $x_b$ is selected from the target domain. Our IMix follows the idea of DACS and applies a ClassMix-based data augmentation for domain adaptation. The differences between the proposed IMix and DACS lie in two aspects. First, we are not using random sampling to select classes instead, we select classes based on the ECS values (Eq. (\ref{eq:ema_ecs})). Second, we do not use a fixed ratio, e.g., $0.5$, but a dynamic schedule to determine the value of the selection ratio. Details about ClassMix and IMix can be found in Algorithm \ref{algo:mix}.

Our IMix is based on the spatiotemporal change of data during training. We want to find a finer structure for this change to make the mixup more focused on bottleneck classes. The intuition is two-fold. First, both domains have increasing performance as the training proceeds, the allocation ratio for mixing regions should incline to the target domain in a gradual manner as our goal is to boost the inference capability on the target domain. However, in the early phase of training, the ratio for source classes should be high as we still want to extract the main knowledge from the source domain. Second, we can empirically find that the performance of some classes is inferior to others during training. The overall performance might be significantly improved if those inferior classes are ameliorated. As we can learn, the dominating domain and the ratio for mixing should be adaptively adjusted along the training process.

\begin{table*} \vspace{-10pt}
\centering
\caption{ Quantitative comparison for the adaptation of GTA-V $\rightarrow$ Cityscapes. \vspace{-8pt}
}\footnotesize
\label{tab:gta}\linespread{1.2} 
\renewcommand{\arraystretch}{1.1}
\setlength{\tabcolsep}{4pt}
\begin{tabular}{l|ccccccccccccccccccc|c}
\hline \hline
Method & \rotatebox[origin=l]{90}{Road} & \rotatebox[origin=l]{90}{S.Walk} & \rotatebox[origin=l]{90}{Build} & \rotatebox[origin=l]{90}{Wall} & \rotatebox[origin=l]{90}{Fence} & \rotatebox[origin=l]{90}{Pole} & \rotatebox[origin=l]{90}{T. Light } & \rotatebox[origin=l]{90}{Sign} & \rotatebox[origin=l]{90}{Veg} & \rotatebox[origin=l]{90}{Terrian} & \rotatebox[origin=l]{90}{Sky} & \rotatebox[origin=l]{90}{Person} & \rotatebox[origin=l]{90}{Rider} & \rotatebox[origin=l]{90}{Car} & \rotatebox[origin=l]{90}{Truck} & \rotatebox[origin=l]{90}{Bus} & \rotatebox[origin=l]{90}{Train} & \rotatebox[origin=l]{90}{MC} & \rotatebox[origin=l]{90}{Bike} & mIoU \\
\hline
APODA \cite{yang2020adversarial} & 85.6 & 32.8 & 79.0 & 29.5 & 25.5 & 26.8 & 34.6 & 19.9 & 83.7 & 40.6 & 77.9 & 59.2 & 28.3 & 84.6 & 34.6 & 49.2 & 8.0 & 32.6 & 39.6 & 45.9\\
PatchAlign \cite{tsai2019domain} & 92.3 & 51.9 & 82.1 & 29.2 & 25.1 & 24.5 & 33.8 & 33.0 & 82.4 & 32.8 & 82.2 & 58.6 & 27.2 & 84.3 & 33.4 & 46.3 & 2.2 & 29.5 & 32.3 & 46.5\\
AdvEnt \cite{vu2019advent} & 89.4 & 33.1 & 81.0 & 26.6 & 26.8 & 27.2 & 33.5 & 24.7 & 83.9 & 36.7 & 78.8 & 58.7 & 30.5 & 84.8 & 38.5 & 44.5 & 1.7 & 31.6 & 32.4 & 45.5\\
CBST \cite{zou2018unsupervised} & 91.8 & 53.5 & 80.5 & 32.7 & 21.0 & 34.0 & 28.9 & 20.4 & 83.9 & 34.2 & 80.9 & 53.1 & 24.0 & 82.7 & 30.3 & 35.9 & 16.0 & 25.9 & 42.8 & 45.9\\
MRKLD-SP \cite{zou2019confidence} & 90.8 & 46.0 & 79.9 & 27.4 & 23.3 & 42.3 & 46.2 & 40.9 & 83.5 & 19.2 & 59.1 & 63.5 & 30.8 & 83.5 & 36.8 & 52.0 & 28.0 & 36.8 & 46.4 & 49.2\\
BDL \cite{li2019bidirectional} & 91.0 & 44.7 & 84.2 & 34.6 & 27.6 & 30.2 & 36.0 & 36.0 & 85.0 & 43.6 & 83.0 & 58.6 & 31.6 & 83.3 & 35.3 & 49.7 & 3.3 & 28.8 & 35.6 & 48.5\\
CADASS \cite{yang2021context} & 91.3 & 46.0 & 84.5 & 34.4 & 29.7 & 32.6 & 35.8 & 36.4 & 84.5 & 43.2 & 83.0 & 60.0 & 32.2 & 83.2 & 35.0 & 46.7 & 0.0 & 33.7 & 42.2 & 49.2\\
MRNet \cite{zheng2019unsupervised} & 89.1 & 23.9 & 82.2 & 19.5 & 20.1 & 33.5 & 42.2 & 39.1 & 85.3 & 33.7 & 76.4 & 60.2 & 33.7 & 86.0 & 36.1 & 43.3 & 5.9 & 22.8 & 30.8 & 45.5\\
R-MRNet \cite{zheng2021rectifying} & 90.4 & 31.2 & 85.1 & 36.9 & 25.6 & 37.5 & 48.8 & 48.5 & 85.3 & 34.8 & 81.1  & 64.4 & 36.8 & 86.3 & 34.9 & 52.2 & 1.7 & 29.0 & 44.6 & 50.3\\
PIT \cite{lv2020cross} & 87.5 & 43.4 & 78.8 & 31.2 & 30.2 & 36.3 & 39.9 & 42.0 & 79.2 & 37.1 & 79.3 & 65.4 & 37.5 & 83.2 & 46.0 & 45.6 & 25.7 & 23.5 & 49.9 & 50.6\\
SIM \cite{wang2020differential} & 90.6 & 44.7 & 84.8 & 34.3 & 28.7 & 31.6 & 35.0 & 37.6 & 84.7 & 43.3 & 85.3 & 57.0 & 31.5 & 83.8 & 42.6 & 48.5 & 1.9 & 30.4 & 39.0 & 49.2\\
FDA \cite{yang2020fda} & 92.5 & 53.3 & 82.4 & 26.5 & 27.6 & 36.4 & 40.6 & 38.9 & 82.3 & 39.8 & 78.0 & 62.6 & 34.4 & 84.9 & 34.1 & 53.1 & 16.9 & 27.7 & 46.4 & 50.45\\
CAG-UDA \cite{zhang2019category} & 90.4 & 51.6 & 83.8 & 34.2 & 27.8 & 38.4 & 25.3 & 48.4 & 85.4 & 38.2 & 78.1 & 58.6 & 34.6 & 84.7 & 21.9 & 42.7 & 41.1 & 29.3 & 37.2 & 50.2\\
IAST \cite{mei2020instance} & 93.8 & 57.8 & 85.1 & 39.5 & 26.7 & 26.2 & 43.1 & 34.7 & 84.9 & 32.9 & 88.0 & 62.6 & 29.0 & 87.3 & 39.2 & 49.6 & 23.2 & 34.7 & 39.6 & 51.5\\
DACS \cite{tranheden2021dacs} & 89.9 & 39.7 & 87.9 & 30.7 & 39.5 & 38.5 & 46.4 & 52.8 & 88.0 & 44.0 & 88.8 & 67.2 & 35.8 & 84.5 & 45.7 & 50.2 & 0.0 & 27.3 & 34.0 & 52.1 \\
CorDA \cite{wang2021domain} & 94.7 & 63.1 & 87.6 & 30.7 & 40.6 & 40.2 & 47.8 & 51.6 & 87.6 & 47.0 & 89.7 & 66.7 & 35.9 & 90.2 & 48.9 & 57.5 & 0.0 & 39.8 & 56.0 & 56.6\\
ProDA \cite{zhang2021prototypical} & 87.8 & 56.0 & 79.7 & 46.3 & \textbf{44.8} & \textbf{45.6} & \textbf{53.5} & \textbf{53.5} & 88.6 & 45.2 & 82.1 & 70.7 & 39.2 & 88.8 & 45.5 & 59.4 & 1.0 & 48.9 & 56.4 & 57.5\\
DAFormer \cite{hoyer2022daformer} & 94.0 & 59.0 & 87.0 & 38.8 & 30.8 & 42.9 & 49.5 & 51.0 & 88.1 & \textbf{48.6} & 89.0 & 69.3 & 39.8 & 91.3 & 72.0 & 69.4 & 48.8 & 52.2 & 61.2 & 62.2\\
HRDA \cite{hoyer2022hrda} & 94.8 & 64.0 & \textbf{88.1} & \textbf{52.7} & 28.2 & 45.5 & 48.4 & 49.2 & \textbf{89.3} & 48.4 & 91.4 & 73.9 & 38.3 & 92.2 & 74.9 & 76.8 & \textbf{62.2} & \textbf{61.5} & 64.1 & 65.4 \\
\hline
IDA (ours) & \textbf{95.4} & \textbf{72.0} & 87.8 & 49.9 & 36.6 & 40.6 & 46.8 & 50.4 & 88.3 & 45.2 & \textbf{92.1} & \textbf{74.2} & \textbf{50.4} & \textbf{92.8} & \textbf{79.2} & \textbf{81.8} & 53.8 & 61.4 & \textbf{64.5} & \textbf{66.5}\\
\hline \hline
\end{tabular} \vspace{-8pt}
\end{table*}

For the convenience of analysis, we propose two concepts, \textit{Source-Select-Target-Follow} (SSTF) and \textit{Target-Select-Source-Follow} (TSSF). The difference between the two is the order of the class selection --- which domain (selecting domain) provides the guaranteed selection of certain classes while the other one (following domain) acts accordingly. The data from the selecting domain is guaranteed to be exposed more during training, thus dominating the knowledge transfer process. 

When selecting classes from one domain, we also need to decide whether well-performing classes or under-performing  classes should be selected. Different types of classes in differing domains have different values. For example, under-performing  classes in the source domain might indicate a strong signal for selection as those classes are bottleneck classes and the ground-truth label of those classes might boost the performance. On the contrary, the under-performing  classes in the target domain might be out of choice as they can contain too much noise. The quality of each class is represented by the corresponding ECS value (Eq. (\ref{eq:ema_ecs})).

\subsubsection{Adaptation Schedule}
\label{sec:schedule}

\begin{figure} 
{
  \centering
    \subfigure[]{\label{fig:basic_func}\includegraphics[width=0.491\linewidth]{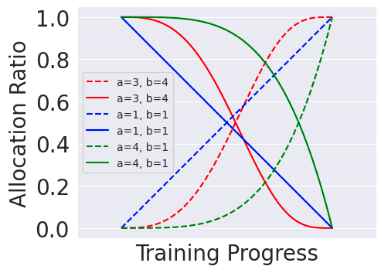}}
    \subfigure[]{\label{fig:schedule_1}\includegraphics[width=0.49\linewidth]{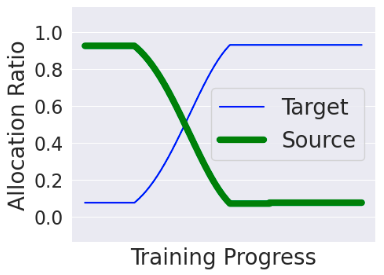}}
  \caption{\small (a) Kumaraswamy Cumulative Density function. (b) The schedule we use for assigning allocation ratio to the dominating domain (shown in a bold curve). In this work, the source domain dominates the allocation.
  }\vspace{-10pt}
\label{fig:schedule}  
}
\end{figure}

To account for the spatial change of the data during training, we propose a dynamic schedule to determine the value of $\eta$ in the function of \textcolor{green!50!black}{\texttt{\textbf{ISample}}} in Algorithm \ref{algo:mix}. The reason for using a dynamic schedule rather than a fixed ratio value is that any single fixed ratio might fail to capture the change throughout the training process. An extreme ratio, e.g., $0.1$ or $0.9$ can lead to a highly imbalanced mixing.
We use the Kumaraswamy Cumulative Density Function (KCDF) as our basic scheduling function. The KCDF has a formulation of $y=1-(1-x^a)^b$. Different KCDFs with different parameters $a$ and $b$ are shown as the solid curve in Fig.~\ref{fig:basic_func}. Another variant of KCDF is expressed by $y=(1-x^a)^b$, we denote this variant as Reversed KCDF (RKCDF), and show different RKCDFs as the dash curves in Fig.~\ref{fig:basic_func}. Based on this basic function, we propose to use a truncated version of the functions to avoid extreme ratio values (see Fig. \ref{fig:schedule_1}).

\section{Experiments}
\label{sec:exp}

\vspace{-3pt}
\subsection{Evaluation Setup} \vspace{-3pt}
\noindent
\textbf{Datasets:}
We test on three datasets.
(1) \textbf{GTA-V} is a synthetic dataset collected in a simulated city environment. This dataset contains $24,966$ synthetic frames with a resolution of $1914\times 1052$. Images are provided with dense semantic annotations of 33 classes.
(2) \textbf{SYNTHIA} is another city-like synthetic dataset that has $9,400$ synthetic images with a resolution of $1280\times 760$. Pixel-level semantic annotations for 13 classes are provided in SYNTHIA.
(3) \textbf{Cityscapes} is a dataset containing $2,975$ training images and $500$ validation images with a resolution of $2048\times 1024$. All images are collected in real European cities.

We perform two Sim2Real adaptations --- one is the adaptation of GTA-V $\rightarrow$ Cityscapes and the other is the adaptation of SYNTHIA $\rightarrow$ Cityscapes. We evaluate segmentation performance
with the standard {\em mean-Intersection-over-Union (mIoU)} 
metric. Evaluations for both adaptation scenarios are conducted on the $500$ validation images in Cityscapes.

\begin{table*}
\centering
\caption{ Quantitative comparison for the adaptation of SYNTHIA $\rightarrow$ Cityscapes. \vspace{-8pt}
}\footnotesize
\label{tab:synthia}
\renewcommand{\arraystretch}{1.1}
\begin{tabular}{l|cccccccccccccccc|c}
\hline \hline
Method & \rotatebox[origin=l]{90}{Road} & \rotatebox[origin=l]{90}{S.Walk} & \rotatebox[origin=l]{90}{Build} & \rotatebox[origin=l]{90}{Wall} & \rotatebox[origin=l]{90}{Fence} & \rotatebox[origin=l]{90}{Pole} & \rotatebox[origin=l]{90}{T. Light } & \rotatebox[origin=l]{90}{Sign} & \rotatebox[origin=l]{90}{Veg} & \rotatebox[origin=l]{90}{Sky} & \rotatebox[origin=l]{90}{Person} & \rotatebox[origin=l]{90}{Rider} & \rotatebox[origin=l]{90}{Car} & \rotatebox[origin=l]{90}{Bus} & \rotatebox[origin=l]{90}{MC} & \rotatebox[origin=l]{90}{Bike} & mIoU \\
\hline
PatchAlign \cite{tsai2019domain} & 82.4 & 38.0 & 78.6 & 8.7 & 0.6 & 26.0 & 3.9 & 11.1 & 75.5 & 84.6 & 53.5 & 21.6 & 71.4 & 32.6 & 19.3 & 31.7 & 40.0\\
AdvEnt \cite{vu2019advent} & 85.6 & 42.2 & 79.7 & 8.7 & 0.4 & 25.9 & 5.4 & 8.1 & 80.4 & 84.1 & 57.9 & 23.8 & 73.3 & 36.4 & 14.2 & 33.0 & 41.2\\
CBST \cite{zou2018unsupervised} & 68.0 & 29.9 & 76.3 & 10.8 & 1.4 & 33.9 & 22.8 & 29.5 & 77.6 & 78.3 & 60.6 & 28.3 & 81.6 & 23.5 & 18.8 & 39.8 & 42.6\\
MRKLD \cite{zou2019confidence} & 67.7 & 32.2 & 73.9 & 10.7 & 1.6 & 37.4 & 22.2 & 31.2 & 80.8 & 80.5 & 60.8 & 29.1 & 82.8 & 25.0 & 19.4 & 45.3 & 43.8\\
MRNet \cite{zheng2019unsupervised} & 82.0 & 36.5 & 80.4 & 4.2 & 0.4 & 33.7 & 18.0 & 13.4 & 81.1 & 80.8 & 61.3 & 21.7 & 84.4 & 32.4 & 14.8 & 45.7 & 43.2 \\
R-MRNet \cite{zheng2021rectifying} & 87.6 & 41.9 & 83.1 & 14.7 & 1.7 & 36.2 & 31.3 & 19.9 & 81.6 & 80.6 & 63.0 & 21.8 & 86.2 & 40.7 & 23.6 & 53.1 & 47.9 \\
PIT \cite{lv2020cross} & 83.1 & 27.6 & 81.5 & 8.9 & 0.3 & 21.8 & 26.4 & 33.8 & 76.4 & 78.8 & 64.2 & 27.6 & 79.6 & 31.2 & 31.0 & 31.3 & 44.0 \\
CAG-UDA \cite{zhang2019category} & 84.7 & 40.8 & 81.7 & 7.8 & 0.0 & 35.1 & 13.3 & 22.7 & 84.5 & 77.6 & 64.2 & 27.8 & 80.9 & 19.7 & 22.7 & 48.3 & 44.5 \\
IAST \cite{mei2020instance} & 81.9 & 41.5 & \textbf{83.3} & 17.7 & 4.6 & 32.3 & 30.9 & 28.8 & 83.4 & 85.0 & 65.5 & 30.8 & \textbf{86.5} & 38.2 & 33.1 & 52.7 & 49.8\\
DACS \cite{tranheden2021dacs} & 80.5 & 25.1 & 81.9 & 21.4 & 2.8 & 37.2 & 22.6 & 23.9 & 83.6 & \textbf{90.7} & 67.6 & 38.3 & 82.9 & 38.9 & 28.4 & 47.5 & 48.3\\
DAFormer \cite{hoyer2022daformer} & 82.3 & 36.9 & 76.1 & 41.8 & \textbf{6.0} & 44.5 & 45.1 & 46.5 & \textbf{85.9} & 82.9 & 68.0 & 44.4 & 84.9 & 47.3 & 49.1 & 57.5 & 56.2 \\
HRDA \cite{hoyer2022hrda} & 83.0 & 43.9 & 76.5 & \textbf{49.8} & 4.3 & \textbf{51.7} & \textbf{55.8} & \textbf{52.8} & 85.2 & 80.0 & 68.2 & 43.0 & 80.7 & 56.7 & \textbf{59.1} & \textbf{59.7} & 59.4 \\
\hline
IDA (ours) & \textbf{88.9} & \textbf{44.2} & 78.2 & 49.1 & 4.9 & 48.6 & 52.3 & 49.3 & 84.9 & 88.2 & \textbf{70.1} & \textbf{47.0} & 85.3 & \textbf{58.2} & 58.7 & 56.9 & \textbf{60.3}  \\
\hline \hline
\end{tabular} \vspace{-8pt}
\end{table*}

\begin{figure*}[t] 
  \centering
   \includegraphics[width=.9\linewidth]{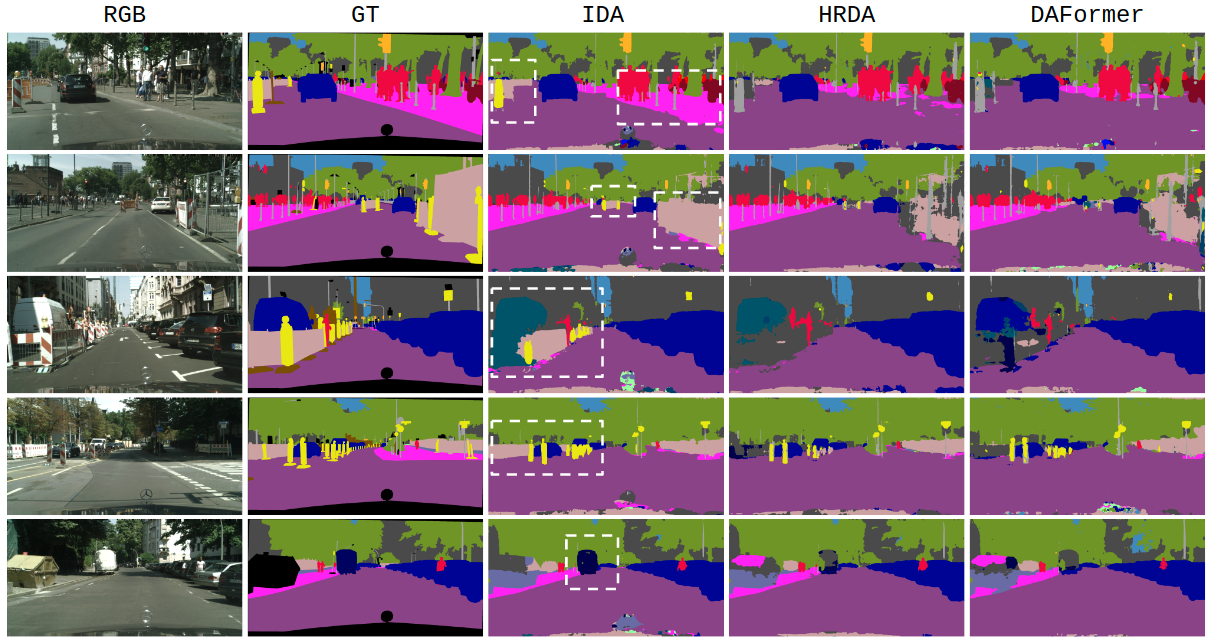} \vspace{-10pt}
   \caption{\small Comparison of different methods on the Cityscapes validation images that show challenging situations. IDA can still maintain a high segmentation quality even when objects are small, ambiguous, and highly irregular, e.g., regions marked by the white dotted boxes.
   \vspace{-10pt}
   }
   \label{fig:city_qualitative}
\end{figure*}

\noindent
\textbf{Implementation Details:}
We base our IDA framework on the self-training framework in HRDA \cite{hoyer2022hrda}. We use a batch size of 1 and set the crop resolution as $952$ due to the limited GPU memory. We compare our proposed IDA with recent SOTA methods 
\cite{hoyer2022daformer} and \cite{hoyer2022hrda}. To make the comparison fair, we also set the same batch size and image resolution for both baselines as ours. More details about network structure and hyperparameters can be found in HRDA \cite{hoyer2022hrda}.

\vspace{-3pt}
\subsection{Comparison}\vspace{-3pt}
We compare our proposed IDA model with the baseline UDA-SS methods both quantitatively and qualitatively. We first consider the adaptation of GTA-V $\rightarrow$ Cityscapes. The quantitative comparison can be seen in Table. \ref{tab:gta}. Our IDA model exhibits the best overall mIoU performance among all the listed methods. IDA outperforms the current UDA-SS SOTA work HRDA \cite{hoyer2022hrda} in the majority of classes (10 out of 19) and shows a considerable advantageous margin of $1.1\%$ mIoU. The IDA model also shows superior performance for some challenging classes, e.g., Person, Rider, Car, Truck, Bus, Bike, etc. This superiority is consistent with our expectation for the IDA model as it especially aims to improve the performance of bottleneck classes during training. 
Note that the results of DAFormer and HRDA in Table~\ref{tab:gta} are different than the reported ones in the original works as we have new hyperparameter settings for a fair comparison with our IDA model. The quantitative comparison for the adaptation of SYNTHIA$\rightarrow$Cityscapes is shown in Table  \ref{tab:synthia}. The IDA model can still outperform the strongest baseline HRDA by a  margin of $0.9\%$ mIoU and show advatanges on challenging classes such as Person, Rider and Bus. 
The effectiveness of our proposed IDA model can also be seen in visual examples in Fig. \ref{fig:city_qualitative}.

\vspace{-3pt}
\subsection{Ablation Studies}\vspace{-3pt}
\label{sec:ablation}

\noindent
\textbf{Fixed selection ratios:}
In IDA we adopt the SSTF-U mixing strategy where we first select the ground-truth regions of source under-performing classes to construct the mixing mask and then the target regions are selected according to the reverse source mask. Throughout the training process, we use a dynamic schedule to determine the value of the ratio for source class selection. 
Thus we conduct the ablation study showing issues with different fixed ratio values under different source class selection strategies, see Table.~\ref{tab:fixed_ratio}.

We show the testing performance on validation data of Cityscapes in the column of \textit{mIoU} and the degradation from the best IDA model in the column of $\delta$. The column of $\Delta$ shows the mean of $\delta$ for each selection with a certain class type. We can see none of the settings in Table.   \ref{tab:fixed_ratio} can achieve a positive $\Delta$. All methods with fixed ratios are degrading with a significant drop. The best result with a fixed ratio (SSTF-U-0.7) still has a gap of $2.3\%$ mIoU compared with the best IDA.

From the value of $\Delta$ in Table.  \ref{tab:fixed_ratio}, we can first validate that the SSTF is the most effective selection method. Based on the SSTF selection, we should give priority to selecting the region of \textit{under-performing} classes in the source domain. Those underperforming classes can be treated as bottleneck classes for both domains, thus they can provide strong supervision to drive the improvement of the overall performance.  An extreme value of the ratio might cause unacceptable damage to the adaptation. The performance with a mild value of the selection ratio, e.g., $0.5$ or $0.7$ can be better than other values, but still worse than the performance with dynamic scheduling. 
Our IDA model is even better than HRDA. The reason for this is we use the dynamic scheduling to balance the bias that we have introduced into the data, thus we are able to extract knowledge from under-performing classes while maintaining the data unbiased. One thing we need to note is that HRDA \cite{hoyer2022hrda} uses a selection ratio of $0.5$, but the performance of HRDA is better than IDA-SSTF-U-0.7. The reason for this drop is the bias we have injected by the selection strategy of SSTF-U  to either well-performing classes or underperforming classes. On the contrary, HRDA uses  random sampling such that the classes of the new data are not biased to any certain types, leading to better results.


\begin{table}
\centering
\caption{Comparison of using different ratio values under different selection strategies. } \vspace{-8pt}
\footnotesize
\label{tab:fixed_ratio}
\begin{tabular}{c|c|c|c|cc|c}
\hline \hline 
& Selection & Class & Ratio & mIoU & $\delta$ ($\uparrow$) & $\Delta$ ($\uparrow$) \\
\hline
0 & \multirow{5}{*}{SSTF} & \multirow{5}{*}{W} & 0.1 & 37.2 & -29.3 & \multirow{5}{*}{-22.5} \\  
1 &  & & 0.3 & 41.4 & -25.1\\ 
2 &  & & 0.5 & 50.7 & -15.8\\ 
3 &  & & 0.7 & 49.2 & -17.3\\ 
4 &  & & 0.9 & 41.3 & -25.2\\ 
\hline
5 & \multirow{5}{*}{SSTF} & \multirow{5}{*}{U} & 0.1 & 40.0 & -16.5 & \multirow{5}{*}{-8.32}\\ 
6 &  & & 0.3 & 45.4 & -11.1\\
7 &  & & 0.5 & 58.2 & -3.3\\ 
8 &  & & 0.7 & 55.2 & -2.3\\ 
9 &  & & 0.9 & 48.1 & -8.4\\ 
\hline
10 & \multirow{5}{*}{TSSF} & \multirow{5}{*}{W} & 0.1 & 50.6 & -15.9 & \multirow{5}{*}{-25.9}\\ 
11 &  & & 0.3 & 47.3 & -19.2\\ 
12 &  & & 0.5 & 38.4 & -28.1\\ 
13 &  & & 0.7 & 35.0 & -31.0\\ 
14 &  & & 0.9 & 31.4 & -35.1\\ 
\hline
15 & \multirow{5}{*}{TSSF} & \multirow{5}{*}{U} & 0.1 & 47.2 & -19.3 & \multirow{5}{*}{-25.4}\\ 
16 &  & & 0.3 & 49.3 & -17.2\\ 
17 &  & & 0.5 & 45.6 & -20.9\\ 
18 &  & & 0.7 & 33.4 & -33.1\\ 
19 &  & & 0.9 & 30.2 & -36.3\\ 
\hline \hline
\end{tabular} 
\end{table}

\noindent
\textbf{Smoothness of the indicator:}
We use the smoothed ECS as the class-level performance indicator. Here we show the necessity of using the smoothed values instead of the raw values. The quantitative comparison is shown in Table.~\ref{tab:smoothness_ablation}. It can be seen that the performance is generally increasing as the smoothness is lifted. The performance with the raw indicator values is the lowest. \begin{wrapfigure}{r}{0.25\textwidth} 
    \centering
    \includegraphics[width=0.25\textwidth]{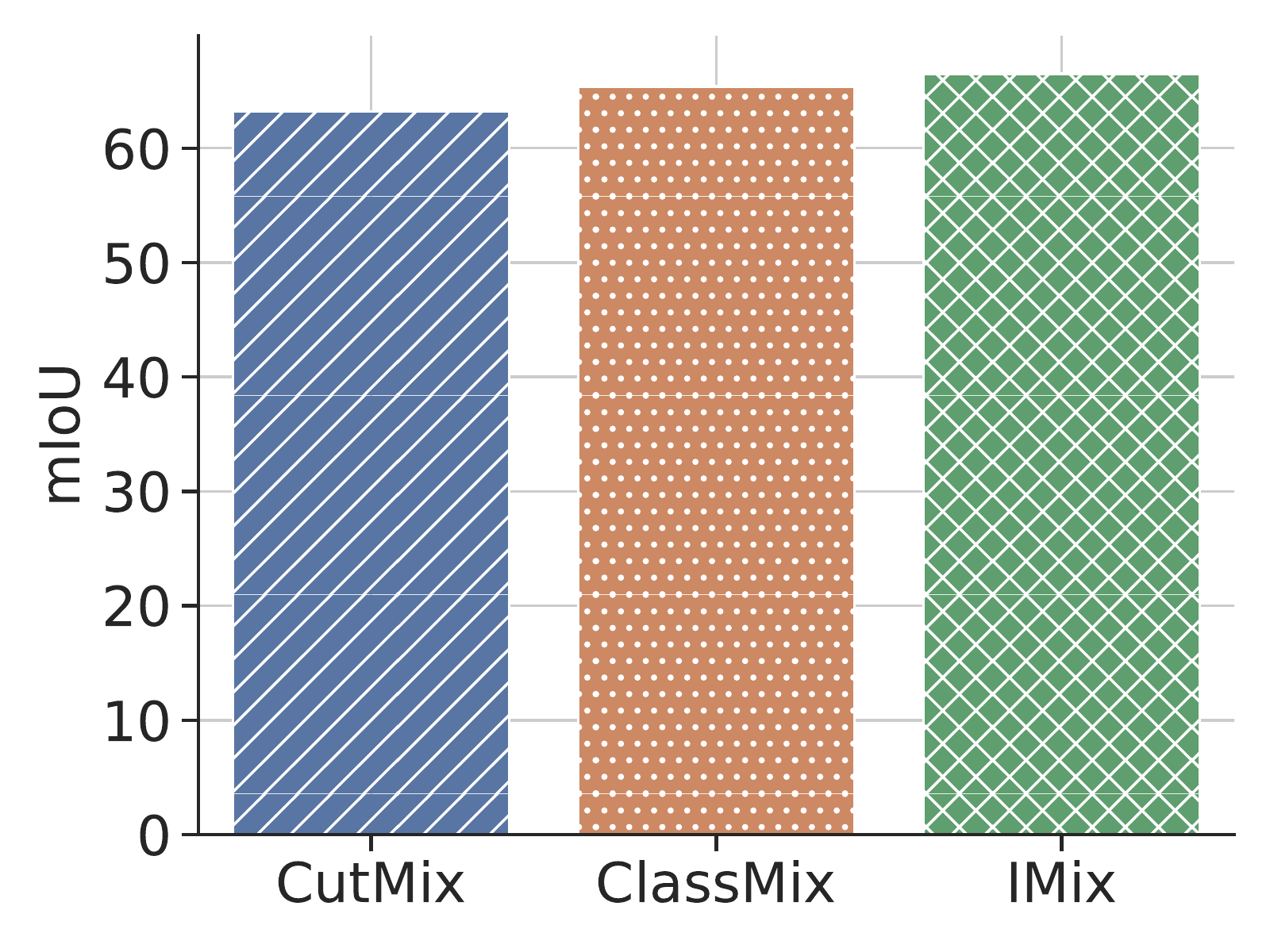}
    \caption{\small Comparison of using different mixups. 
   }
   \label{fig:imix} 
\end{wrapfigure}
~~
The reason for this trend is the smoothed indicator is more stable than the raw values of the indicator which may change significantly among iterations, causing instability in the selection of classes and possibly a large distribution shift 
during training. 

\begin{table}[t]
\centering
\caption{Comparison of using different smoothness for ECS. }\vspace{-8pt}
\small
\label{tab:smoothness_ablation}
\setlength{\tabcolsep}{3.5pt}
\begin{tabular}{c|cccccccc}
\hline \hline 
Smoothness weight & 0 & 0.1 & 0.3 & 0.5 & 0.7 & 0.9 & 0.999  \\
\hline
mIoU & 62.1 & 61.5 & 60.8 & 63.2 & 64.9 & 66.2 & 66.5\\
\hline \hline
\end{tabular} 
\end{table}



\noindent 
\textbf{Different mixups:} In our work we propose a new mixup, IMix, for augmenting data in the IDA model. We compare the performance of the adaptation with different mixups, which is shown in Fig. \ref{fig:imix}. Compared with previous region-based mixups that use random sampling to generate new mixed data, our IMix considers dynamic changes in the data from the two domains. By capturing the fine structure of the adaptation, our IMix achieves the best performance among all listed mixups.

\section{Conclusion} 
We present a principled model, Informed Domain Adaptation (IDA), for the un-supervised domain adaptive semantic segmentation. Our proposed IDA model is a self-training framework that exploits the obscured informativeness of data to improve the learning efficiency. To achieve this, 
we propose a new mixup technique, IMix, that bridges the source and target domains according to the training progress defined by an expected confidence assessment.
We also propose a novel dynamic adaptation schedule which can adaptively adjust the mixing ratio for different domains.
Extensive evaluations on popular datasets reveal that the IDA outperforms the SOTA model with a remarkable margin.


\bibliographystyle{unsrt}
\bibliography{ref}

\end{document}